\documentclass[pdflatex,sn-vancouver-num, iicol]{sn-jnl}

\usepackage{graphicx}%
\usepackage{multirow}%
\usepackage{amsmath,amssymb,amsfonts}%
\usepackage{amsthm}%
\usepackage{mathrsfs}%
\usepackage[title]{appendix}%
\usepackage{xcolor}%
\usepackage{textcomp}%
\usepackage{manyfoot}%
\usepackage{booktabs}%
\usepackage{algorithm}%
\usepackage{algorithmicx}%
\usepackage{algpseudocode}%
\usepackage{listings}%
\usepackage{breakurl}   
\usepackage{amssymb}

\theoremstyle{thmstyleone}%
\theoremstyle{thmstyletwo}%

\theoremstyle{thmstylethree}%

\begin{document}

\title[Article Title]{FBS Model-based Maintenance Record Accumulation for Failure-Cause Inference in Manufacturing Systems}

\author*[1]{\fnm{Takuma} \sur{Fujiu}}\email{takumafujiu@g.ecc.u-tokyo.ac.jp}
\author[1]{\fnm{Sho} \sur{Okazaki}}
\author[2]{\fnm{Kohei} \sur{Kaminishi}}

\author[3]{\fnm{Yuji} \sur{Nakata}}
\author[3]{\fnm{Shota} \sur{Hamamoto}}
\author[3]{\fnm{Kenshin} \sur{Yokose}}

\author[2]{\fnm{Tatsunori} \sur{Hara}}
\author[2]{\fnm{Yasushi} \sur{Umeda}}
\author[2]{\fnm{Jun} \sur{Ota}}

\affil[1]{\orgdiv{Department of Precision Engineering, School of Engineering}, \orgname{The University of Tokyo}, \orgaddress{\street{7-3-1 Hongo}, \city{Bunkyo-ku}, \state{Tokyo} \postcode{113-8656}, \country{Japan}}}

\affil[2]{\orgdiv{Research into Artifacts Center for Engineering (RACE), School of Engineering}, \orgname{The University of Tokyo}, \orgaddress{\street{7-3-1 Hongo}, \city{Bunkyo-ku}, \state{Tokyo} \postcode{113-8656}, \country{Japan}}}

\affil[3]{\orgname{DENSO CORPORATION}, \orgaddress{\street{1-1 Showa-cho}, \city{Kariya}, \state{Aichi} \postcode{448-8661}, \country{Japan}}}

\abstract{
    In manufacturing systems, identifying the causes of failures is crucial for maintaining and improving production efficiency.
    In knowledge-based failure-cause inference, it is important that the knowledge base (1) explicitly structures knowledge about the target system and about failures, and (2) contains sufficiently long causal chains of failures.
    In this study, we constructed Diagnostic Knowledge Ontology and proposed a Function-Behavior-Structure (FBS) model-based maintenance-record accumulation method based on it.
    Failure-cause inference using the maintenance records accumulated by the proposed method showed better agreement with the set of candidate causes enumerated by experts, especially in difficult cases where the number of related cases is small and the vocabulary used differs.
    In the future, it will be necessary to develop inference methods tailored to these maintenance records, build a user interface, and carry out validation on larger and more diverse systems.
    Additionally, this approach leverages the understanding and knowledge of the target in the design phase to support knowledge accumulation and problem solving during the maintenance phase, and it is expected to become a foundation for knowledge sharing across the entire engineering chain in the future.
}


\maketitle

\section{Introduction}\label{sec:introduction}
In the manufacturing industry, maintaining high production efficiency is important.
However, manufacturing systems do not always operate flawlessly, and failures occur frequently.
These breakdown losses are one of the major factors that reduce production efficiency and can severely disrupt overall production planning.
Consequently, when failures occur in manufacturing systems, it is vital to swiftly identify the causes and implement corrective measures.

This task of identifying the causes of failures has traditionally been carried out by experts.
Experts observe the failure that has occurred, infer potential causes based on their knowledge and experience, and then verify the plausibility of these potential causes by inspecting the actual equipment, thereby identifying the true cause of the failure.
However, the shortage of experienced experts has made it difficult for them to respond to every failure.
As a result, non-experts are often required to perform initial troubleshooting.
This situation has created a growing need to support non-experts in identifying the causes of failures.

As a method to support failure cause identification, knowledge-based fault diagnosis is commonly used.
This approach builds a knowledge base from experts’ knowledge and past failure analyses, and uses it to infer the causes of failures.
Although various knowledge-based fault diagnosis methods have been proposed, their adoption in actual production sites remains limited.
A major challenge in identifying the causes of failures in manufacturing systems lies in the need for both deep and shallow knowledge in knowledge-based fault diagnosis\cite{zhou2015}.
Here, deep knowledge refers to knowledge about the structure and function of the target manufacturing system, while shallow knowledge refers to empirical knowledge about failures and their causal relationships.
For example, the knowledge that a robot's mechanical chuck is driven by a servo is considered deep knowledge, while the idea that, in assembly processes, failures such as misalignment are often caused by the workpiece gripping operation is considered shallow knowledge.

First, with regard to deep knowledge, it is necessary to understand the various elements within the target system and their interrelationships.
Manufacturing systems are complex, consisting of numerous pieces of equipment with diverse configurations and layouts.
Therefore, in knowledge-based fault diagnosis, it is especially important to determine what aspects of the system should be represented in the model used for inference.
Specifically, key relationships that must be captured in the model are hierarchical, realization, and sequential relationships.
Regarding hierarchical relationships, Zhou et al. (2015) stated that in complex systems, failures propagate along hierarchical structures, making it necessary to analyze the system hierarchically \cite{zhou2015}.
AIAG \& VDA (2019), who proposed a standardized method for FMEA (Failure Mode and Effects Analysis), also suggested performing structure analysis and function analysis prior to failure analysis, based on a hierarchical understanding of system structure and function\cite{AIAG2019}.
Another important type of relationship is the realization relationship among function, behavior, and structure, which is similar to hierarchical relationships.
In the field of design engineering, Gero (1990) proposed the FBS (Function-Behaviour-Structure) model, stating that function is realized through behavior, and behavior is realized through structure \cite{gero1990}.
From the perspective of system failures, the realization relationship implies that a functional failure can result from a behavioral failure, which can result from a structural failure.
Regarding sequential relationships, Okazaki et al. (2023) pointed out that in fault diagnosis for manufacturing systems, the sequential relationship between processes is also important, since a failure occurring in an earlier process can affect subsequent processes.
In this way, the important relationships that need to be captured as system knowledge when analyzing failures in manufacturing systems are complex in nature.
Therefore, in knowledge-based fault diagnosis, where the causes of new failures are inferred from a knowledge base consisting of past failure analysis results, it is crucial that these relationships are explicitly organized and represented in the accumulated analysis results.

Shallow knowledge, that is knowledge related to failures, is equally important, which must be captured with a causal chain of sufficient length.
For example, the 'Five Whys' method proposed by Toyota indicates that identifying the root cause of a manufacturing failure typically requires tracing the causal chain through at least five layers \cite{barsalou2023}.
To repeatedly follow these causal relationships while considering the hierarchical, realization, and sequential relationships mentioned above, it is essential that the referenced information supports causal reasoning along chains of sufficient length.

From these considerations, two essential requirements emerge for a knowledge base in knowledge-based fault diagnosis: (1) an explicitly structured linkage between deep and shallow knowledge, and (2) causal chains of failures with sufficient length.
Crucially, it is not just the individual requirements but their simultaneous fulfillment that is essential.

This study proposes a FBS model-based maintenance-record accumulation method that simultaneously satisfies these two essential requirements for knowledge-based failure-cause inference in manufacturing systems.
Specifically, deep knowledge is organized by integrating the functional hierarchy of AIAG \& VDA with FBS (Function-Behavior-Structure), and within this framework, failure events and causal relationships are systematically linked and accumulated.
This provides a foundation for reasoning that considers hierarchical, realization, and sequential relationships within manufacturing systems, thereby satisfying both requirements (1) and (2).
Details are provided in Section 3.

The contributions of this paper are threefold:
(i) an ontology that satisfies the requirements of linking deep and shallow knowledge and providing causal chains of sufficient length,
(ii) an operational scheme to accumulate maintenance records on the FBS model, and
(iii) a bridge between the design and maintenance phases.

The remainder of this paper is structured as follows.
Section~\ref{sec:related_work} positions this study within the context of related work.
Section~\ref{sec:method} describes the proposed method in detail.
Section~\ref{sec:experimental_setup} outlines the experimental setup, using a LEGO car assembly line as a case study.
Section~\ref{sec:result_discussion} discusses the experimental results.
Finally, Section~\ref{sec:conclusion} presents our conclusions and directions for future research.

\section{Related Work}\label{sec:related_work}
\subsection{FMEA as an Explicitly Structured Knowledge Source}
Failure Mode and Effects Analysis (FMEA) is a representative reliability- and quality-analysis method employed across a wide range of industries, and it is also a widely used knowledge source for knowledge-based fault diagnosis\cite{wu2021}.
Conducted during the design phase of a manufacturing system by a team of multiple experts, FMEA exhaustively enumerates potential failures, organizing their causes, effects, and countermeasures.
Industry standards such as IEC 60812:2018 \cite{IEC60812_2018} and the AIAG \& VDA FMEA Handbook \cite{AIAG2019} codify both the tabular format and the execution workflow for FMEA.
Thanks to these structured, table-based templates and the clearly prescribed procedural steps, FMEA achieves an explicitly structured linkage between deep knowledge, documented in the FMEA items ``Structure'' or ``Function'', and shallow knowledge, documented in ``Failure Mode,'' ``Cause,'' and ``Effects''.

In the use of FMEA, various approaches have been proposed, including constructing ontologies from FMEA descriptions to enable search and inference, and using Case-Based Reasoning (CBR).
Ontology is described by Studer et al. (1998) as "a formal, explicit specification of a shared conceptualisation" \cite{studer1998}.
CBR is a reasoning method that solves new problems based on similarities with past cases. It enables problem solving by repeating a cycle of retrieval, reuse, revision, and retention \cite{kolodner1992, prentzas2007}.
FMEA ontologies have been developed for domains such as automobiles \cite{rehman2016}, wind turbines \cite{zhou2015}, and pressure vessels \cite{hodkiewicz2021}.
In these studies, elements related to failures were retrieved from the FMEA data by performing OWL-DL reasoning or querying with SPARQL (SPARQL Protocol and RDF Query Language) \cite{seaborne2008}.
Arifin et al. (2023) proposed a method for failure-cause inference using CBR, targeting an automotive airbag system.
In this method, FMEA items are treated as cases, and similarities across various attributes are calculated to perform case-based reasoning \cite{arifin2023}.
Okazaki et al. linked FMEA items with process sequence models of manufacturing systems described in SysML\cite{friedenthal2014}, managed them using an ontology, and applied CBR to them\cite{okazaki2023}.

However, FMEA-based approaches have a shortcoming with respect to the length of causal chains.
Okazaki et al. (2023) evaluated their method by comparing its inference results with failure-cause candidates listed by experts for several assumed failures, and noted that the FMEA-based inferences were inferior in terms of granularity\cite{okazaki2023}.
This indicates that inferences based solely on FMEA lack sufficient length in their causal chains.
This is likely due to the fact that, while FMEA is a standardized and systematic reliability analysis method, it is fundamentally oriented toward comprehensive coverage rather than length of causal chains.

Consequently, FMEA satisfies one of the two goals pursued in this study: (1) an explicitly structured linkage between deep and shallow knowledge, but it does not fulfill the other goal, which is (2) causal chains of failures with sufficient length.

\subsection{Maintenance Records as a Knowledge Source with Sufficient Causal-Chain Length}
Maintenance records are logs of the maintenance activities actually carried out during the maintenance phase of manufacturing systems.
Typically, in addition to descriptions of the occurred failure and its cause, information such as countermeasures is also included.
They offer a sufficiently high degree of granularity, yet they lack any consistent textual style or content, which makes information retrieval and reuse difficult.
Hershowitz et al. (2024) point out that technicians tend to describe failure events in their own individual ways using unstructured, free-form text\cite{hershowitz2024}.
Consequently, at the present stage, we have little choice but to rely on natural-language-processing (NLP) techniques to make practical use of maintenance records.
Recently, numerous studies have aimed to extract causal relations from maintenance records for Root Cause Analysis, typically employing Transformer-based deep-learning models \cite{hershowitz2024,xu2022,zhou2022}.
For similarity search within maintenance records, researchers first apply NLP techniques such as Named Entity Recognition (NER) or Part-of-Speech (POS) tagging, or generate semantic embeddings with BERT, and then perform reasoning with CBR or knowledge graphs \cite{devaney2005,xu2025,naqvi2022,naqvi2024}.

However, these techniques address only shallow knowledge and provide little linkage to knowledge of the target system, or deep knowledge.
To connect shallow knowledge sources like maintenance records with deep knowledge, the necessary information and relationships must be explicitly documented.
That is, items and their contents must be systematically managed at the time of data collection.
Some approaches have been proposed, such as those by Fujiu et al. (2024) and Fernández del Amo (2024), which manage maintenance records from the point of accumulation using predefined ontology classes and properties\cite{fujiu2024,fernandez2024}.
However, they have not yet addressed the organization of deep knowledge using ontologies or the integration with maintenance records.

\subsection{Ontological Modeling of Systems for Fault Diagnosis}
How to model deep knowledge, that is, knowledge about the target system, is critically important in fault diagnosis, and has been discussed in many studies.
In ontological modeling of manufacturing systems, as noted in Section \ref{sec:introduction}, it is necessary to capture (a) hierarchical relationships, (b) realization relationships, and (c) sequential relationships in order to understand the propagation of failures.

Hierarchical analysis of the target system is not uncommon and has been adopted in many studies \cite{zhou2015, xu2018}.
Within these studies, hierarchies such as Device, System, (sub)Assembly, and Part are defined.
However, a key issue is that no method has been provided for how to interpret or recognize these hierarchies in the context of fault diagnosis.

Realization relationships among function, behavior, and structure are framed by the FBS model in design engineering, where functions are realized through behaviors, and behaviors are realized through structures\cite{gero1990,gero2007}.
In the context of FBS, the function of an object is defined as its teleology (``what the object is for''), while behavior is defined as ``what the object does'', and structure is defined as ``what the object consists of''\cite{gero2007}.
Younus et al. (2024) proposed a method that combines FBS and failure analysis (FMEA) in the product-design phase, but this approach only considers realization relationships and does not sufficiently address how to achieve causal chains of adequate length in fault diagnosis, such as what kinds of hierarchies should be established\cite{younus2024}.

Okazaki et al. (2023) described the sequence of processes (specifically, instances of the \texttt{function} class in their ontology) using diagrams created with SysML, and constructed a process order model by integrating these diagrams.
As repeatedly mentioned, they regard the lack of detail in the inference results as a problem, which is considered to stem from the fact that their modeling does not take hierarchical relationships into account.

Although various types of modeling have been explored in previous studies, there is currently no method that successfully expresses the three relationships that must be defined on the deep knowledge side in order to satisfy the first requirement of this study, which concerns the explicit and structured linkage between deep and shallow knowledge.

\subsection{Contributions of This Study}
The contributions of this work are threefold:
\begin{enumerate}
    \item \textbf{Diagnostic Knowledge Ontology that bridges deep and shallow knowledge}:
    We propose an ontology that serves as a foundation for modeling the target system and recording failures, in order to satisfy two important requirements in failure-cause inference using maintenance records: (1) an explicitly structured linkage between deep and shallow knowledge, and (2) causal chains of failures with sufficient length.
    In this ontology, deep knowledge about the target system is represented using three types of relationships, namely hierarchical, realization, and sequential relationships.
    Descriptions of failures, which constitute shallow knowledge, are linked to this deep knowledge together with the causal relationships among the failures.
    \item \textbf{Operational scheme to accumulate maintenance records on the FBS model}:
    This study presents a concrete operational workflow for accumulating maintenance records based on the FBS model of manufacturing systems.
    First, at the start of operation of the manufacturing system, the FBS model of the target system is constructed as instances of the classes related to deep knowledge in the Diagnostic Knowledge Ontology.
    During operation, each maintenance event is recorded as an instance of the classes related to shallow knowledge and linked to the FBS model.
    This approach enables continuous and structured accumulation of maintenance knowledge in alignment with the proposed ontology.
    \item \textbf{Bridge between the design and maintenance phases}:
    The structuring of deep knowledge in this study is based on design-phase methodologies such as FMEA and FBS, and can be seen as an approach that bridges the design and maintenance phases.
    This unification enables knowledge continuity throughout the entire manufacturing system life cycle, grounded in design-oriented thinking.
    In the future, this structure is expected to serve as a foundation for comprehensive knowledge sharing across various phases of the system life cycle, not only in fault diagnosis but also in prognostics and health management, as well as system reconfiguration.
\end{enumerate}

\section{Proposed Method}\label{sec:method}
\subsection{Overview of the Proposed Framework}

\begin{figure*}[t]
    \begin{center}
    \includegraphics[width=15cm]{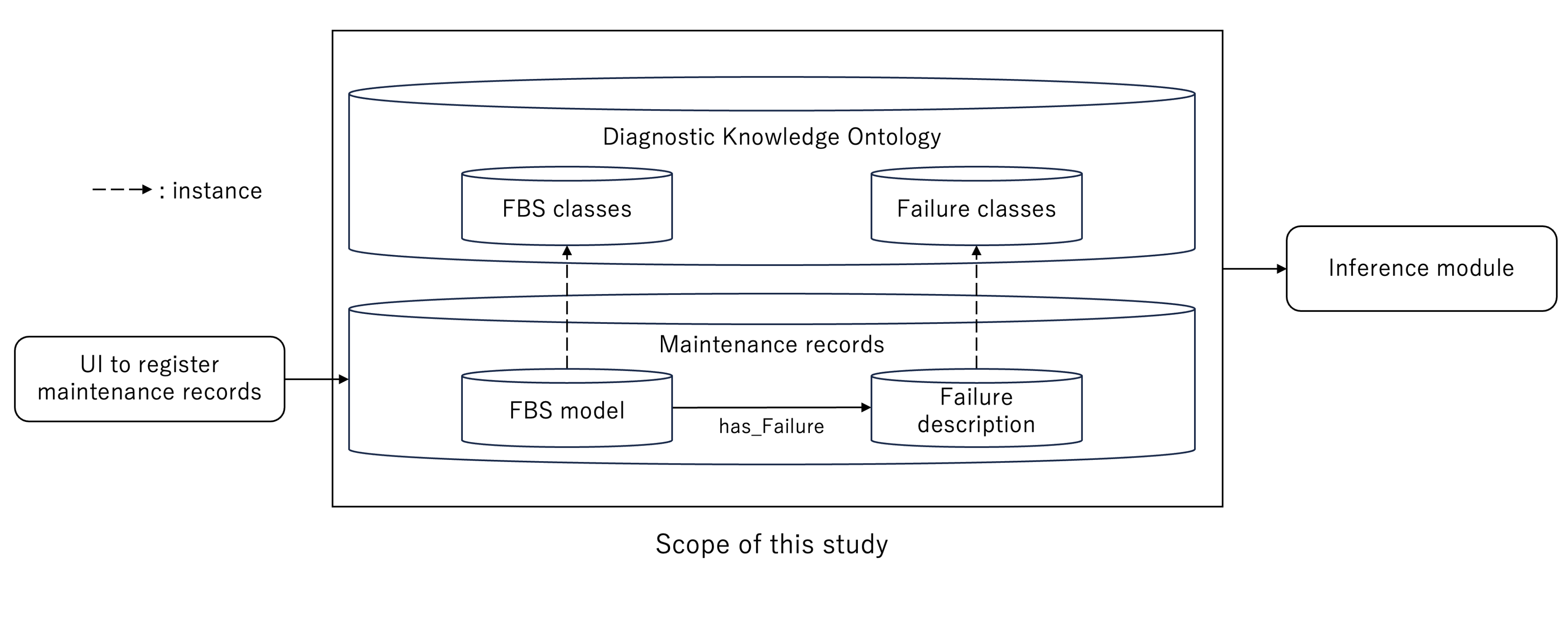}
    \caption{Overview of the proposed framework}
    \label{fig:overall_framework}
    \end{center}
\end{figure*}

The overall framework of the proposed method for accumulating maintenance records is shown in Fig.~\ref{fig:overall_framework}.
This accumulation of maintenance records is intended to serve as a knowledge base for failure-cause inference in manufacturing systems.
The records are accumulated via a user interface outside the scope of this study and are similarly used by an inference module that is also beyond the study’s scope.
At the core of the maintenance-record accumulation, which falls within the scope of this study, is the Diagnostic Knowledge Ontology, described in detail in Subsection~\ref{subsec:dkonto}.
This ontology integrates deep knowledge about the target manufacturing system with shallow knowledge about failures.
Specifically, the system’s functions, behaviours, and structures are represented as instances of deep-knowledge classes defined in the ontology.
Because these classes are organized according to the FBS paradigm, the collection of such instances is referred to as the system’s FBS model.
Analyses of observed failures are attached to this FBS model and stored as instances of the ontology’s shallow-knowledge classes.
In this study, the FBS model of the manufacturing system and the associated failure analyses taken together are referred to as the maintenance records.

The remainder of this section is organized as follows.
Subsection~\ref{subsec:dkonto} describes the classes and properties defined by the Diagnostic Knowledge Ontology.
Subsection~\ref{subsec:accumulation} explains how the maintenance records are actually recorded and stored.

\subsection{Diagnostic Knowledge Ontology}\label{subsec:dkonto}

This subsection describes the ontology used to store maintenance records.  
An ontology can serve as a backbone for structuring domain knowledge and for keeping maintenance data well organized.  
We refer to the ontology proposed in this study as the Diagnostic Knowledge Ontology.

\begin{figure*}[h]
    \begin{center}
    \includegraphics[width=16cm]{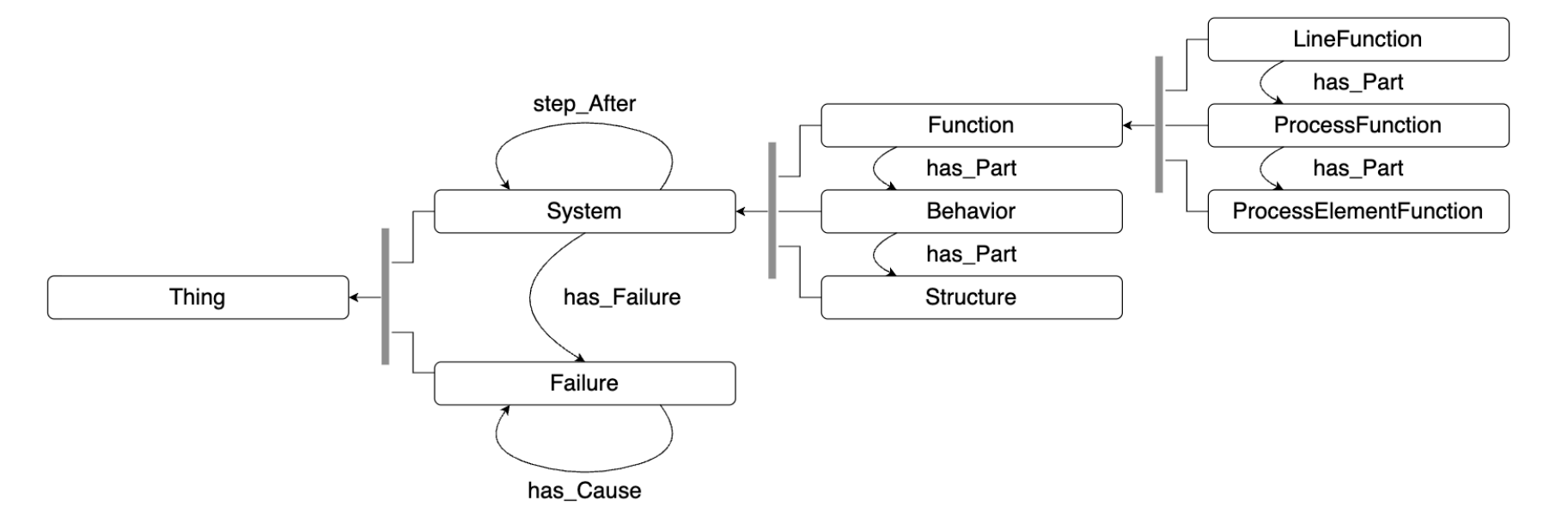}
    \caption{Diagnosis knowledge ontology}
    \label{fig:ontology}
    \end{center}
\end{figure*}

The main classes and properties of the ontology are shown in Fig.~\ref{fig:ontology}.  
At the top level, the ontology contains two primary classes: \texttt{System} and \texttt{Failure}.  
The \texttt{System} class has three subclasses—\texttt{Function}, \texttt{Behavior}, and \texttt{Structure}.
It follows the Function-Behaviour-Structure (FBS) paradigm, which asserts that a system’s functions are realised by behaviours, and behaviours are realised by structures.  

The \texttt{Function} class is further divided into the subclasses \texttt{LineFunction}, \texttt{ProcessFunction}, and \texttt{ProcessElementFunction}, which stand in a hierarchical relationship.
This hierarchy follows the Function Analysis step in the AIAG \& VDA (2019) FMEA methodology.
AIAG \& VDA (2019) first perform a hierarchical analysis of structure and then define the functions achieved at each structural level.  
Table~\ref{tab:process_hierarchy} lists the three structural levels defined by AIAG \& VDA (2019).

\begin{table*}[h]
    \caption{Hierarchy of Structure proposed by AIAG \& VDA (2019)\cite{AIAG2019}}
    \centering
    \begin{tabular}{|p{5cm}|p{10cm}|}
        \hline
        \textbf{Level of structure} & \textbf{Description} \\
        \hline
        \textbf{Process Item} & The highest level of the structure tree or process flow diagram and PFMEA. The end result of all of the successfully completed Process Steps. \\
        \hline
        \textbf{Process Step} & A manufacturing operation or station. The function of the process step describes the resulting product features produced at the station. \\
        \hline
        \textbf{Process Work Element} & The lowest level of the process flow or structure tree. The function of the process work element reflects its contribution to the process step to create the process. \\
        \hline
    \end{tabular}
    \label{tab:process_hierarchy}
\end{table*}

In practice, however, a manufacturing system is composed of a variety of elements arranged in line-specific layouts, so there is no single, uniform structure.
Therefore, experts' interpretation of the relationship between structure and function varies, for example, when one robot assembles several parts, someone call each part-assembly operation a process, whereas others call the entire set of robot motions a process.
On actual production sites, multiple terms such as ``station,'' ``module,'' and ``process'' are often used interchangeably without clear distinctions.
For this reason, it is difficult to define the hierarchy of structures before defining that of functions in the manufacturing system.
To overcome this problem, it is helpful to reinterpret system functions in terms of their impact on products.
In the proposed Diagnostic Knowledge Ontology, we adopt the unified view summarised in Table \ref{tab:line_function}, which enables consistent understanding across different production lines.

\begin{table*}[h]
    \caption{The definition of functional hierarchy in manufacturing systems proposed in this study, along with its correspondence to the definition by AIAG \& VDA (2019) \cite{AIAG2019}.}
    \centering
    \begin{tabular}{|p{3cm}|p{8cm}|p{4cm}|}
        \hline
        \textbf{Level of Function} & \textbf{Description} & \textbf{Correspondence with AIAG \& VDA (2019)~\cite{AIAG2019}} \\
        \hline
        \textbf{Line Function} & The role that the line should fulfill for the next line or end users & Function of Process Item \\
        \hline
        \textbf{Process Function} & The role of making changes to the product in terms of workmanship quality. ``Workmanship'' refers to the conditions that the processed workpiece must meet to become a finished product. In general manufacturing methods, workmanship includes factors such as the processing position, correct part number, and absence of damage. & Function of Process Step \\
        \hline
        \textbf{Process Element Function} & The effect that process elements (5M) have on the product in order to achieve the function of the process: the degree of detail in changing the product’s shape, position and orientation, and information. & Function of Process Work Element \\
        \hline
    \end{tabular}
    \label{tab:line_function}
\end{table*}

An important point here is that even the lowest-level function \texttt{ProcessElementFunction} still produces an effect on the work-in-process product.
However, actual maintenance records discuss not only failures in teleological functions that affect products, but also failures in the behaviors and structures of the manufacturing systems.
Therefore, in addition to the three functional levels (\texttt{LineFunction}, \texttt{ProcessFunction}, and \texttt{ProcessElementFunction}), the Diagnostic Knowledge Ontology introduces the \texttt{Behavior} that realises the lowest-level function, \texttt{ProcessElementFunction}, and the \texttt{Structure} that realises that \texttt{Behavior}.
In total, the ontology thus represents a manufacturing system with five levels.

We adopt the generic property \texttt{has\_Part} as an overarching relation that covers both (i) the realisation chain linking \texttt{Function}~$\rightarrow$~\texttt{Behavior}~$\rightarrow$~\texttt{Structure} and (ii) the hierarchical decomposition of functional subclasses.
Because failures can propagate along either path, using a single, unified relation simplifies causal reasoning in the ontology.
In practice, in identifying failure causes on production sites, engineers shift perspectives from function to behavior and then to structure.
This is because teleological functions tend to be more general than behaviors or structures.
For example, an assembly function can be realized by either a mechanical chuck or a vacuum unit.
This order is therefore effective for narrowing down possible causes.

In practice, in identifying failure causes on production sites, engineers typically start from functions, move to the associated behaviors, and finally inspect the underlying structures.
Because teleological functions tend to be more general or shared than behaviors or structures, for example, an assembly function can be realised either by a mechanical chuck or by a vacuum gripper, so descending from functions to behaviors and then to structures is an effective way to narrow potential causes.
Reflecting this practice, we organize the ontology into five successive levels, from \texttt{LineFunction} down to \texttt{Structure}, connected exclusively by \texttt{has\_Part}.

Order relations among the same-level subclasses of \texttt{System} are represented by the property \texttt{step\_After}.

The \texttt{Failure} class represents a failure that occurs in the manufacturing system.  
It has an attribute \texttt{FailureCategory}, which classifies the failure, for example, as motion-related, mechanism/structure-related, or accuracy-related.  
Causal links between failures are captured by the properties \texttt{has\_Cause} and \texttt{has\_Effect}.  
The property \texttt{has\_Failure} connects a failure analysis to the FBS model of the target system, linking instances of the \texttt{Failure} class to instances of the \texttt{System} class.

By designing the ontology in this way, a foundation is established for accumulating maintenance records that meet the two requirements of this study: (1) an explicitly structured linkage between deep and shallow knowledge, and (2) causal chains of failures with sufficient length.

\subsection{Maintenance-Record Documentation and Storage}\label{subsec:accumulation}
In this section, we explain how maintenance records are actually described and stored as instances of the Diagnostic Knowledge Ontology introduced in the previous section.
The overall procedure consists of two main steps: (1) constructing the FBS model of the target system, and (2) Storing the maintenance records themselves in accordance with that model.

\subsubsection{FBS Model Construction}
The FBS model of the target system, shown in the left side of Fig. \ref{fig:maintenance_records}, is constructed at system commissioning, prior to the storage of any maintenance records. 
Information about the target system is registered as instances of the subclasses of \texttt{System} defined in the Diagnostic Knowledge Ontology.  
By using the ontology’s classes and properties that capture the system’s hierarchical (\texttt{has\_Part}) and sequential (\texttt{step\_After}) relationships, all system data required for failure-cause inference are saved in a unified, machine-interpretable form.  
Consequently, when inference is performed, elements of the target system can be compared systematically, and the inference engine can determine which previously stored case knowledge is relevant.

\subsubsection{Maintenance Record Storage}
\begin{figure*}[t]
    \begin{center}
    \includegraphics[width=16cm]{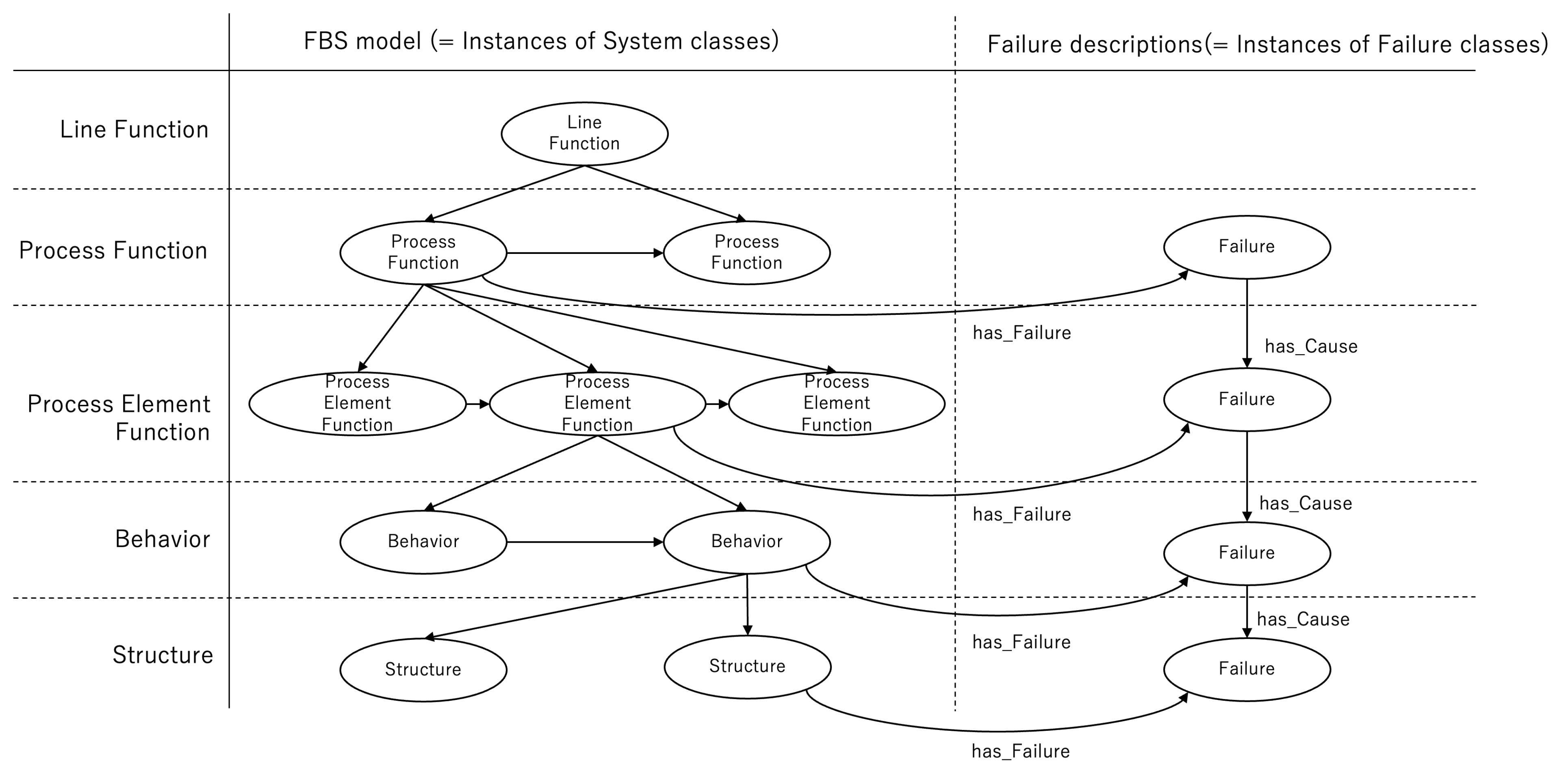}
    \caption{At the system commissioning stage, the FBS model of the target system is constructed. After the system begins operation, descriptions of failures are accumulated in a manner that links them to this model.}
    \label{fig:maintenance_records}
    \end{center}
\end{figure*}

In the maintenance-record storage step, each observed failure is documented in terms of two kinds of relations:  (i) the causal links among failures (\texttt{has\_Cause}) and (ii) the links between a failure and the system function in which it occurred (\texttt{has\_Failure}, shown in the right side of Fig. \ref{fig:maintenance_records}).  

Concretely, the node where the failure arose is selected from those in the target system’s FBS model and is connected to the corresponding failure instance via the property \texttt{has\_Failure}.  
Causal relationships between failure instances are captured by the property \texttt{has\_Cause}.  

With these links in place, each failure case is stored together with an explicit indication of where in the target system it occurred, enabling precise case-retrieval during failure-cause inference.

\section{Experimental Setup}\label{sec:experimental_setup}
In this experiment, we compared two types of knowledge bases for failure-cause inference in a manufacturing line: (1) maintenance records linked with the FBS model (proposed method), and (2) maintenance records excluding FBS information (baseline).
For inference, we employed the chunk generation and RAG (Retrieval-Augmented Generation) method proposed by Bahr et al. (2025)\cite{bahr2025}.
Two assumed failures were introduced into the target system, the LEGO car assembly line: (1) “assembly misalignment,” (2) “roof misalignment in chuck.”
The evaluation was conducted by comparing the inference outputs with the “ground truth” cause-list provided by experts.
The performance metrics used were $Precision@n$ and $Recall@n$.
This allowed us to investigate the impact of the knowledge base with FBS model proposed in this study on inference accuracy.

\subsection{Knowledge-Base Configurations and Inference Engine}
\begin{figure*}[h]
    \begin{center}
    \includegraphics[width=16cm]{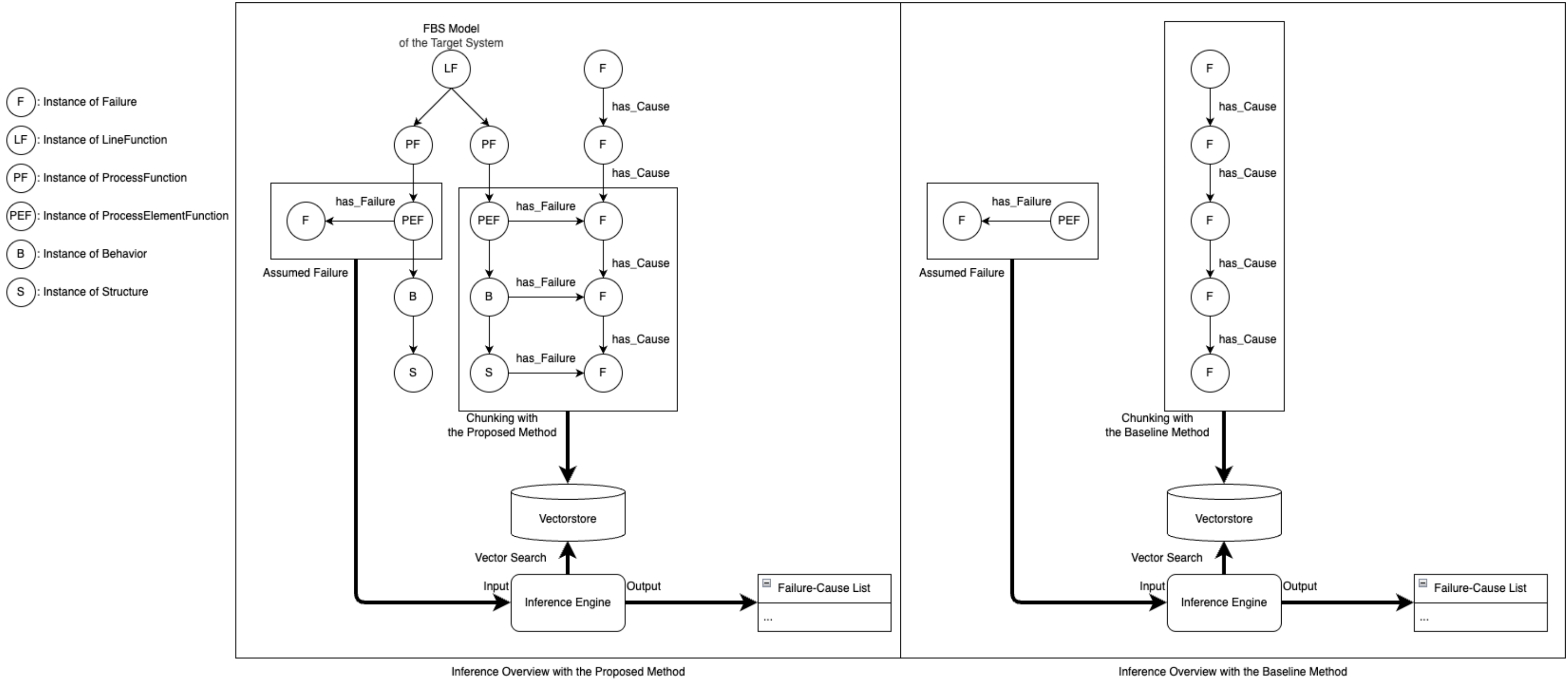}
    \caption{Overview of inference in the experiment and comparison of chunking methods between the proposed method and the baseline. In the proposed method, chunks are limited to the hierarchical level where the assumed failure occurred and below, and they include both failure and system information. In contrast, the baseline method does not consider hierarchy and treats a whole causal chain of failures as a single chunk.}
    \label{fig:method_comparison}
    \end{center}
\end{figure*}

In this experiment, we evaluated the performance of failure-cause inference based on maintenance records stored using two different methods, as summarized in Table~\ref{tab:method_comparison} and Fig.~\ref{fig:method_comparison}.

The maintenance records used in the proposed method contain deep knowledge representing information about the target system, shallow knowledge representing failures and their causal relationships, and the relationships between the deep and shallow knowledge, all structured using the Diagnostic Knowledge Ontology.
In contrast, the baseline method uses ablated maintenance records in which the FBS model and the relationships between the system and failures are removed, shown in Table~\ref{tab:method_comparison}.
This ablation is intended to reproduce the current situation in which deep knowledge is not recorded in maintenance records.
This comparison can test whether the explicit and structured information in the Diagnostic Knowledge Ontology improves inference.

For the inference engine, we adopted the chunking and RAG method proposed by Bahr et al. (2025)\cite{bahr2025}.
In this approach, a subgraph from the knowledge graph is first converted into a chunk, which is then used for RAG.
RAG uses an embedding model to convert chunks of information into vector representations in a semantic space.
These vectors capture the meaning of the content.
When a query is given, RAG searches this semantic space to retrieve the chunks that are relevant to the input.
As shown in Fig. \ref{fig:method_comparison}, in the proposed method, chunking is first performed by identifying the failure node at the same hierarchical level with input failures, and then forming a chunk that includes the failure nodes causally related to it, as well as the FBS-model nodes connected to them.
In contrast, in the baseline, each maintenance records' item forms a single chunk regardless of hierarchy.
The vector embedding is then computed using OpenAI’s \texttt{text-embedding-3-small} model.

\begin{table*}[h]
    \caption{Comparison of the maintenance records between the proposed method and the baseline. In the proposed method, the maintenance records used are fully supported by the Diagnostic Knowledge Ontology. In contrast, the baseline uses ablated maintenance records in which the FBS model, which represents information about the target system, and the relationships between the model and failure descriptions are removed.}
    \centering
    \begin{tabular}{|p{2.5cm}|p{6cm}|p{6cm}|}
        \hline
        & \textbf{Proposed Method} & \textbf{Baseline} \\
        \hline
        \textbf{Description}
        & Maintenance records stored as instances of the Diagnostic Knowledge Ontology, in which each failure description is linked to the FBS model of the target system
        & FBS model and all links to it were removed, leaving only failure descriptions and their causal relationships \\
        \hline
        Instances of \texttt{Failure} class      & \checkmark & \checkmark \\
        \hline
        Instances of \texttt{System} subclasses  & \checkmark & -- \\
        \hline
        \texttt{has\_Cause}                      & \checkmark & \checkmark \\
        \hline
        \texttt{has\_Failure}                   & \checkmark & -- \\
        \hline
    \end{tabular}
    \label{tab:method_comparison}
\end{table*}

\subsection{Experimental Environment and Data}
In this experiment, we targeted a learning factory, which is an automated assembly line for education, training, and research purposes.
Learning factories simulate real production chains and serve as neutral testing grounds, making their validated results equivalent to those from actual manufacturing lines\cite{abele2017}.
The learning factory used here constitutes one segment of a LEGO-car assembly line.

\begin{figure}[h]
    \centering
    \includegraphics[width=7cm]{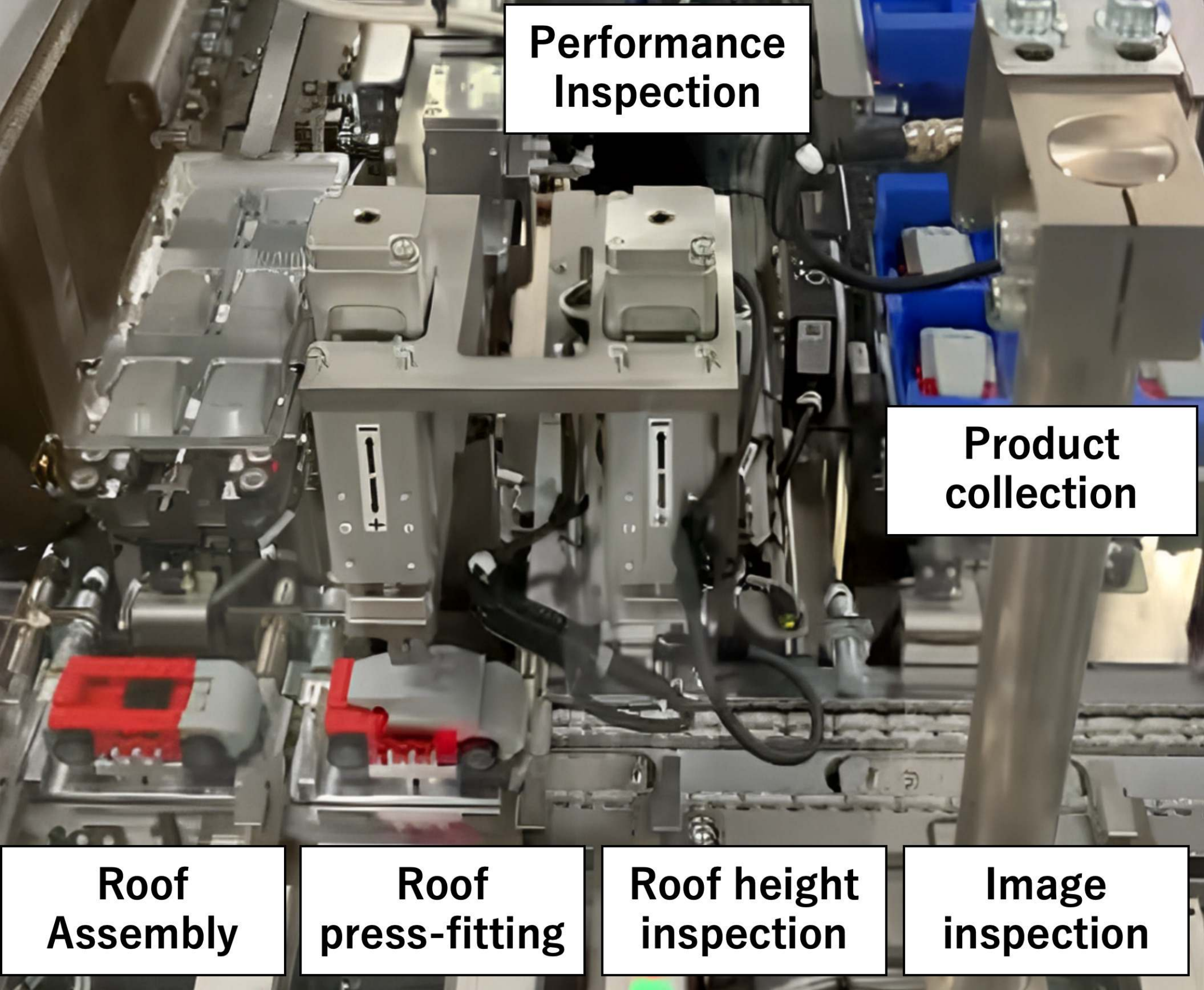}
    \caption{The manufacturing system used in this experiment is a part of a LEGO-car assembly line.  
    It consists of six processes: roof assembly, roof press-fitting, roof-height inspection, image inspection, performance inspection, and product collection.}
    \label{fig:lf_mc3}
\end{figure}

As illustrated in Fig.~\ref{fig:lf_mc3}, the line comprises six processes: roof assembly, roof press-fitting, roof-height inspection, image inspection, performance inspection, and product collection.
A single robot located at the center of these processes performs the roof transfer in the roof assembly, transportation to the performance inspection station, and product collection.
Roof press-fitting and roof-height inspection are performed by separate cylinders.
The workpieces are placed on pallets and transported through these processes by a conveyor.

The maintenance records used for this experiment comprise 168 entries describing failures that either actually occurred on the line or could realistically occur.
These records were compiled mainly by two expert technicians.

Two assumed failures were selected for the inference experiment:
Failure (1): ``assembly misalignment'' occurring in the process function ``assemble roof onto workpiece, ''
and Failure (2): ``roof misalignment inside chuck during transportation'' occurring in the process-element function ``chuck the roof.''

These failures were selected because they occur at different levels, with Failure (1) at the process level and Failure (2) at the process-element level.
This difference is reflected in the amount and wording of related maintenance data.
For Failure~(1), a sufficient number of relevant maintenance records are available in this experiment, and the phrase supplied as input is almost identical to the wording stored in those records.  
In contrast, Failure~(2) is associated with only a few maintenance records, and even those describe the failure in wording that differs markedly from the input description.

Consequently, Failure~(1) is expected to be comparatively easy for the inference engine to diagnose, whereas Failure~(2) is regarded as a more challenging case.

\subsection{Evaluation Method}
In this experiment, the possible failure causes listed by experts for the assumed failures are regarded as the ``ground truth,'' and the inference results are evaluated based on their degree of match with the ground truth.
The ground truth is based on data collected in the studies of Okazaki et al. (2023) and Fujiu et al. (2024), utilizing interview results conducted with two experts who have decades of experience in the design and setup of assembly systems\cite{okazaki2023,fujiu2024}.
For Assumed Failure~(1), there are 35 ground truth items, and for Assumed Failure (2), there are 8.

The accuracy of failure-cause inference is evaluated using the following two metrics \cite{fujiu2024}:
\begin{itemize}
    \item $Precision@n$ : The proportion of outputs within the top $n$ that match the ground truth.
    \item $Recall@n$: The proportion of the ground truth items that is covered by the top $n$ outputs.
\end{itemize}
It is an important point that, among the two evaluation metrics, $Recall@n$ is the more important one.
This is because if the inference results do not include the true failure cause, investigations will not lead to its identification, which can significantly reduce production efficiency.
In this experiment, we assume that the ground truth provided by experts includes the true failure cause, so it is important to cover more of the ground truth, in other words, $Recall@n$ is more important.

For each assumed failure, the maximum number of outputs ($n$) is limited to the number of ground truth items listed by the experts.

\subsection{Implementation}
The ontology and knowledge graph were implemented mainly with Prot\'eg\'e, Owlready2, and Neo4j.  
Prot\'eg\'e provides an intuitive graphical interface for building ontologies\cite{mark2015}, whereas Owlready2 is a Python package for ontology-oriented programming\cite{lamy2021}.  
Neo4j, one of the most widely used graph databases, was employed to store the knowledge graph\cite{neo4j2019}.

Using these tools, we constructed the FBS model of the target line and stored the associated maintenance records.  
The FBS model was represented as a knowledge graph comprising 165 nodes and 220 edges.  
The maintenance records were then linked to this model and stored as 1176 additional nodes.  

Note that a dedicated user interface that would allow shop-floor personnel to perform these tasks intuitively has not yet been implemented.

\section{Results and Discussion}\label{sec:result_discussion}
\begin{figure*}[t]
\centering
\includegraphics[width=0.9\textwidth]{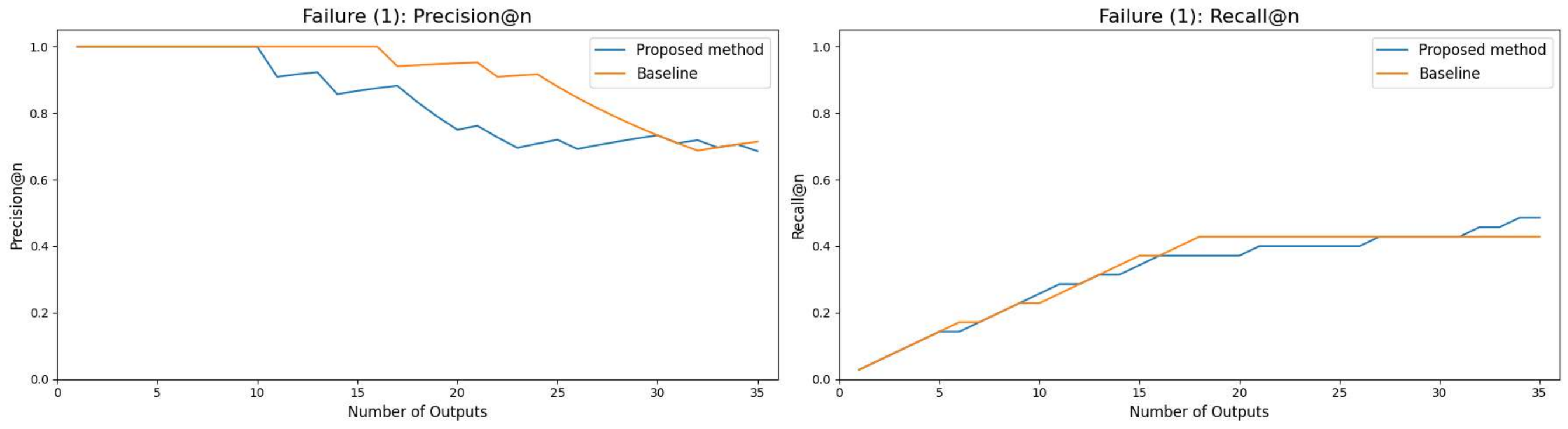}
\caption{Results for Assumed Failure (1): Assembly misalignment.}
\label{fig:failure1}
\end{figure*}

\begin{figure*}[t]
\centering
\includegraphics[width=0.9\textwidth]{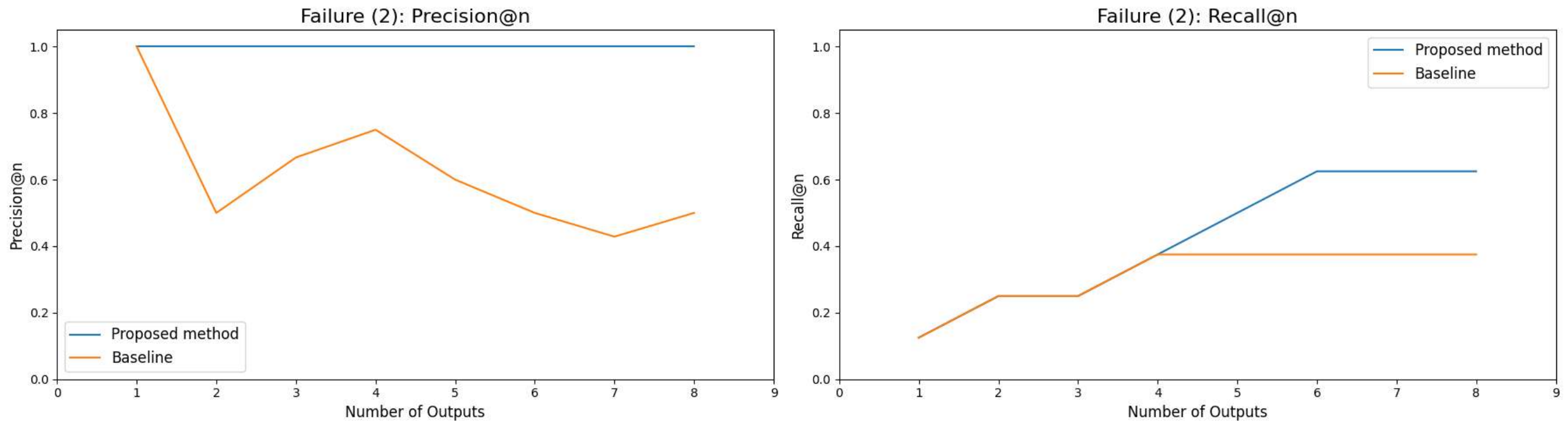}
\caption{Results for Assumed Failure (2): Roof misalignment inside chuck during transportation.}
\label{fig:failure2}
\end{figure*}

The $Precision@n$ and $Recall@n$ for Assumed Failure 1 in Fig.~\ref{fig:failure1} (assembly misalignment) and Assumed Failure 2 in Fig.~\ref{fig:failure2} (roof misalignment inside the chuck) are shown below.

Assumed Failure 1 represents an easy case, because a large number of relevant maintenance records have been accumulated and the same wording is used almost uniformly.
Under this condition, the performance of the baseline and the proposed method was generally comparable.
At $n = 35$, the proposed method slightly underperformed the baseline in terms of $Precision@35$, but outperformed it in terms of $Recall@35$, which is more important from a practical standpoint.

By contrast, for Assumed Failure 2, where only a few exact maintenance records exist and the wording does not match, the proposed method surpassed the baseline by a large margin across both metrics.
For example, at $n = 8$, the proposed method achieved $Precision@8 = 1.0$ and $Recall@8 = 0.625$, while the baseline only reached $Precision@8 = 0.50$ and $Recall@8 = 0.375$.
This indicates that, for the proposed method, all outputs matched the ground truth, covering 5 out of 8 ground truth instances, with 3 of the 8 outputs being duplicates of already identified correct outputs.

Two main factors account for this improvement with the proposed method.
First, because the FBS model simultaneously incorporates not only the failure descriptions but also the information of the target manufacturing system, similarity could be computed appropriately even in the presence of wording variations.
In the inference engine used in this experiment, it is presumed that the information from those nodes was also embedded into a single vector, which was then reflected in the search results.
Second, the FBS model explicitly retains hierarchy information, therefore, chunk generation and search could be restricted to failure nodes within the same hierarchical level, effectively reducing the search space.

On the other hand, for certain values of $n$, we observed regions where $Precision@n$ of the proposed method fell below that of the baseline even though $Recall@n$ was similar.
It should be noted that the inference engine employed in this experiment has not been fine-tuned for the proposed FBS-model-based maintenance-record framework, therefore, the following weaknesses may partly stem from this mismatch rather than from the proposed method itself.
This decline of $Precision@n$ is presumed to result from the proposed method uniformly treating surrounding nodes and generating embeddings as a single chunk with the inference engine used in this study.
In searches aimed at failure in the assembly process, some failure nodes from the press-fitting process were included.
This likely occurred because, in the target system of this experiment, assembly and press-fitting include the same lower-level process elements such as palette-positioning and robot-positioning, which caused the chunk embedding vectors to become similar.
To suppress this behavior, it would be effective to generate embeddings on a node-by-node basis rather than embedding an entire subgraph into a single chunk, and to apply weights according to the hierarchy or distance between nodes, as an inference engine fine-tuned for maintenance records stored with the proposed method.

\section{Conclusion}\label{sec:conclusion}
This study proposes an FBS model-based schema for describing and accumulating maintenance records for failure-cause inference in manufacturing systems, which simultaneously guarantees explicit structural linkage between deep knowledge (knowledge of the target system) and shallow knowledge (knowledge about failures) as well as sufficiently long causal chains of failures.
Based on the proposed Diagnostic Knowledge Ontology, an FBS model of the manufacturing system is constructed, and by attaching each failure analysis to the corresponding nodes of the FBS model, maintenance records are accumulated in an integrated form that unifies the target system’s function, behavior, and structure with the causal relations among failure events.

Experimental evaluation showed that the knowledge graph built with the proposed method improves failure-cause inference.
In particular, for the difficult test case, where only a small number of relevant maintenance records existed and their lexical expressions differed, the proposed method markedly surpassed the baseline in precision and recall, confirming that the ontology-driven integration of deep and shallow knowledge mitigates the effects of vocabulary mismatch and data scarcity.

These findings underscore the importance of reusing design-phase system definitions during the maintenance phase.
The proposed ontology-based knowledge accumulation method is expected not only to shorten troubleshooting time during maintenance, but also to serve as a knowledge-sharing foundation for the entire engineering chain.

Future work will address:
(1) development of an inference engine with a node-level embeddings;
(2) a user-friendly interface for shop-floor engineers; and
(3) validation on larger, more diverse manufacturing systems to confirm generalisability.

\bibliography{sn-article}

\section*{Statements and Declarations}
\paragraph*{Funding}
This research was conducted in Research into Artifacts Center for Engineering (RACE), School of Engineering, The University of Tokyo.  
It was partly supported by a joint research agreement with DENSO CORPORATION.  
No additional external funding was received for this study.
\paragraph*{Competing Interests}
The authors declare that they have no competing financial or non-financial interests.  
\paragraph*{Author Contributions}
Takuma Fujiu: Conceptualization, methodology, software implementation, ontology construction, experiments, data analysis, visualization, and original draft preparation.\\
Sho Okazaki: Major contributor to manuscript revision, experimental design, and in-depth discussions on research content.\\
Kohei Kaminishi: Supervision of the project team. Contributed to research discussions and manuscript revision.\\
Yuji Nakata, Shota Hamamoto, Kenshin Yokose: Provided expert knowledge of actual manufacturing sites, supplied experimental data, and participated as expert collaborators in the experiments.\\
Tatsunori Hara, Yasushi Umeda, Jun Ota: Academic Supervision.


\end{document}